\documentclass{article}

 \usepackage{graphicx}
 \usepackage{subfigure}
\usepackage{algorithmic}
\usepackage{algorithm}
\usepackage[utf8]{inputenc} 
\usepackage[T1]{fontenc}    
\usepackage{hyperref}       
\usepackage{url}            
\usepackage{booktabs}       
\usepackage{amsfonts}       
\usepackage{nicefrac}       
\usepackage{microtype}      
\usepackage{xcolor}         
\usepackage{stfloats}
\usepackage{float}  
\usepackage{wrapfig}
\usepackage{wrapfig,lipsum,booktabs}
\usepackage{amsmath}
\usepackage{adjustbox}
\usepackage{multirow}

\newcommand{\hv}[0]{\ensuremath{\boldsymbol{h}} }
\newcommand{\iv}[0]{\ensuremath{\boldsymbol{i}} }

\newcommand{\given}{\,|\,}

\usepackage{hyperref}

\usepackage{hyperref}

\usepackage{amsmath}
\usepackage{amssymb}
\usepackage{mathtools}
\usepackage{amsthm}

\usepackage[capitalize,noabbrev]{cleveref}

\theoremstyle{plain}

\theoremstyle{definition}

\theoremstyle{remark}

    \usepackage[preprint]{neurips_2024}

\usepackage[utf8]{inputenc} 
\usepackage[T1]{fontenc}    
\usepackage{hyperref}       
\usepackage{url}            
\usepackage{booktabs}       
\usepackage{amsfonts}       
\usepackage{nicefrac}       
\usepackage{microtype}      
\usepackage{xcolor}         

\title{Beyond Spectral Decomposition: Bayesian Contrastive Learning and
its Non-negative Formulation via Factor Analysis}

\author{\hspace{1mm}Zhibin Duan\thanks{Equal Contribution.} \\
Xidian University\\
\texttt{xd\_zhibin@163.com} \\
	\And
\hspace{1mm}Tiansheng Wen\footnotemark[1] \\
Xidian University\\
\texttt{neilwen987@stu.xidian.edu.cn} \\
	\And
\hspace{1mm}Yifei Wang \\
MIT CSAIL\\
\texttt{yifei\_w@mit.edu} \\
	\And
\hspace{1mm}Chen Zhu \\
Xidian University\\
\texttt{22009100527@stu.xidian.edu.cn} \\
	\And
\hspace{1mm}Bo Chen\thanks{Corresponding author} \\
Xidian University\\
\texttt{bchen@mail.xidian.edu.com} \\
\And
\hspace{1mm}Mingyuan Zhou \\
The University of Texas at Austin\\
\texttt{mingyuan.zhou@mccombs.utexas.edu} \\
}

\begin{document}

\maketitle

\begin{abstract}
Factor analysis, often regarded as a Bayesian variant of matrix factorization, offers superior capabilities in capturing uncertainty, modeling complex dependencies, and ensuring robustness.
As the deep learning era arrives, factor analysis is receiving less and less attention due to their limited expressive ability. 
On the contrary, contrastive learning has emerged as a potent technique with demonstrated efficacy in unsupervised representational learning.
While the two methods are different paradigms, recent theoretical analysis has revealed the mathematical equivalence between contrastive learning and matrix factorization, providing a potential possibility for factor analysis combined with contrastive learning.
Motivated by the interconnectedness of contrastive learning, matrix factorization, and factor analysis, this paper introduces a novel Contrastive Factor Analysis framework, aiming to leverage factor analysis's advantageous properties within the realm of contrastive learning. 
To further leverage the interpretability properties of non-negative factor analysis, which can learn  disentangled representations, contrastive factor analysis is extended to a non-negative version.
Finally, extensive experimental validation showcases the efficacy of the proposed contrastive (non-negative)  factor analysis methodology across multiple key properties, including expressiveness, robustness, interpretability, and accurate uncertainty estimation.
\end{abstract}

\section{Introduction}
Before the era of deep learning, factor analysis (FA) \citep{chen2013deep} had received widespread attention and achieved impressive results.
Generally, FA is a statistical method used to identify underlying factors or latent variables that explain
patterns of correlations among observed variables. 
With its stochastic latent variables, FA often enjoys good properties for modeling complex dependence and enhancing robustness \citep{zhou2016augmentable}. 
Meanwhile, its non-negative extensions, such as Poisson factor analysis \citep{zhou2012beta}, can not only learn meaningful latent representation from high-dimension observation unsupervised but also can discover interpretable semantic concepts 
 \citep{blei2003latent, lee1999learning}. 
With these appealing properties, FA and non-negative FA have been widely used in many applications, including text analysis \citep{zhou2012beta} image representation learning \citep{chen2013deep}. 

With the deep learning era arriving \citep{lecun2015deep}, FA is receiving less and less attention due to its limited expressiveness. 
Correspondingly, recent years have witnessed a surge of interest in self-supervised learning, such as contrastive learning (CL) \citep{oord2018representation, chen2020simple}, owing to its remarkable capacity to learn high-quality representations unsupervised, leading to notable advancements in various fields \citep{you2020graph, gao2021simcse}.
Generally, CL methodologies aim to extract discriminative features by training models to distinguish positive and negative samples. 
The distinction between FA and CL lies primarily in the latter's ability to harness effective training objectives \citep{ye2019unsupervised} and expressive deep neural networks \citep{he2015deep}.

Given that FA and CL are regarded as two distinct research fields, integrating CL to enhance FA or vice versa might appear challenging. Fortunately, recent advancements in the theoretical underpinnings have provided insights into this integration.  Specifically, \cite{haochen2021provable} leverages spectral contrastive loss to derive the mathematical equivalent of CL and matrix factorization (MF) \citep{eckart1936approximation}.
Considering the relationship between MF and FA \citep{nakajima2011theoretical}, the integration of FA with CL becomes feasible. 
Beyond feasibility, merging FA and CL offers numerous attractive potential benefits. 
In general, from the perspective of FA, enhancing FA with CL can alleviate its limitation in expressiveness, leading to richer and more informative representations \citep{bengio2014representation}.
From the perspective of CL, considering FA's capacity to capture complex dependencies and its stochastic latent variables, incorporating CL with FA not only enhances performance but also facilitates improved uncertainty estimation and interpretability \citep{fan2020bayesian, higgins2017beta}.

\begin{figure*}[!t]
\flushleft
\quad\quad
\includegraphics[width=13.5cm]{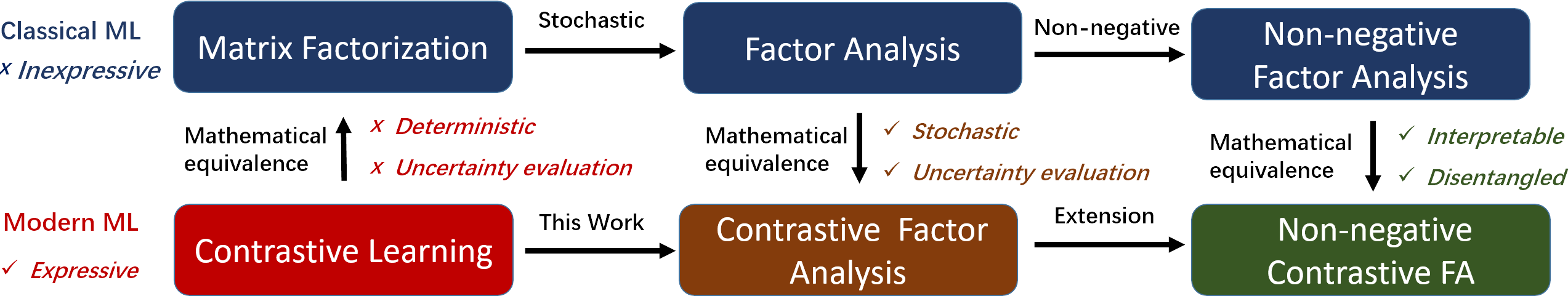} 
\caption{ Relationship between different learning paradigms discussed in this work.}
\label{fig_motivation}
\vspace{-4mm}
\end{figure*} 
Drawing inspiration from the mathematical equivalence established between MF, CL, 
and FA, as illustrated in Figure.~\ref{fig_motivation}, in this paper, we try to develop a Contrastive Factor Analysis (CFA) framework. 
Unlike FA, which directly models observation data, CFA follows CL methods to model the relation matrix, which represents the relationship between positive and negative samples.
More specifically, in CFA, the relation matrix is factorized into Gaussian latent variables. 
Considering the target matrix is non-negative, the CL can also be modified as a contrastive non-negative FA, which factorizes the relation matrix into gamma latent variables.
Compared with CFA, its non-negative extension can enjoy good properties for learning disentangled representation \citep{bengio2014representation, wang2024nonnegative}. 
To approximate the posterior distribution of the latent variables, we further develop a variational inference network.
While it's straightforward to employ the Gaussian distribution to appropriate the posterior of Gaussian latent variables,  considering the non-reparameterizability of the gamma distribution, we instead utilize the Weibull distribution to approximate the posterior of gamma latent variables \cite{zhang2018whai}. 
Finally, both models' parameters can be effectively trained within a variational inference framework \citep{kingma2013auto} with reparameterize technique.

To assess the properties of contrastive factor analysis, we conducted a series of experiments targeting various aspects. 
These properties encompass the attainment of more expressive representations, robustness against out-of-distribution data, interpretability, and uncertainty evaluation. 
{The main contributions of this paper can be summarized as follows:} 
\begin{itemize}
\item {
Drawing inspiration from the interplay between contrastive learning, matrix factorization, and factor analysis, we introduce a novel contrastive (non-negative) factor analysis framework. 
}
\item {To approximate the posterior distribution of latent variables, we propose two variational inference networks: Gaussian and Weibull variational inference network.}

\item{ Extensive experimental validation showcases the efficacy of the proposed contrastive factor analysis methodology across multiple key properties, including expressiveness, robustness, interpretability, and accurate uncertainty estimation.}
\end{itemize}

\section{Preliminary on contrastive learning}
\begin{figure*}[h]
\centering
\subfigure[CL]{
\includegraphics[height=2.3cm]{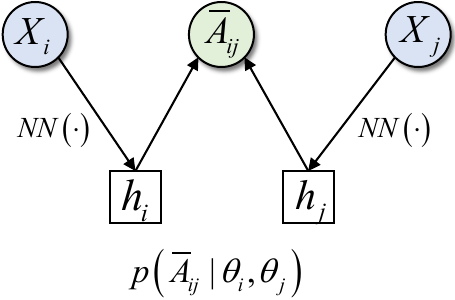}
\label{fig_gbn}
}\quad\quad\quad
\subfigure[Generative model]{
\includegraphics[height=2.3cm]{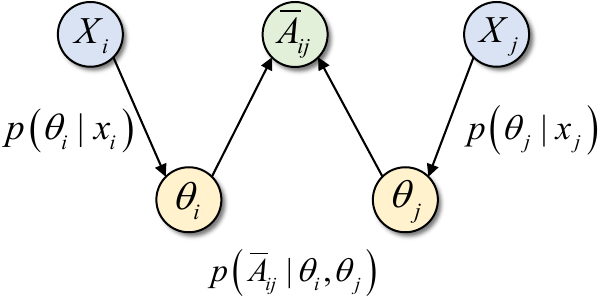}
\label{fig_gm}
}\quad\quad\quad
\subfigure[VI]{
\includegraphics[height=2.3cm]{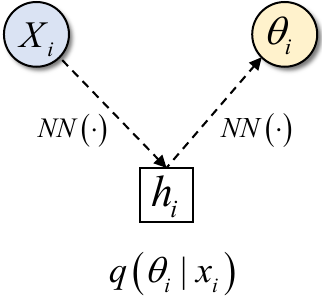}
\label{fig_vi}
} \vspace{-2mm}
\caption{The graphical model of \textit{~\ref{fig_gbn}:} contrastive learning (CL)
\textit{\ref{fig_gm}:} the generative model of contrative factor analysis;  \textit{\ref{fig_vi}:} variational inference (VI) of contrative factor analysis.
Circles are stochastic variables, and squares are deterministic variables. 
 } \label{fig_inference_net}
\vspace{-2mm}
\end{figure*}

In this section, we briefly review the standard contrastive learning modules and the relationship between contrastive learning and matrix factorization.
\paragraph{Contrastive learning.} 
In representation learning, the aim is to learn an encoder function  $f: {\mathbb{R}^D} \to {\mathbb{R}^d}$ that maps high-dimensional input data  $x \in {\mathbb{R}^D}$ to low-dimensional representations $z \in {\mathbb{R}^d}$. 
To perform contrastive learning, we start by building pairs of positive samples $\left( {x,{x_ + }} \right)$ by augmenting the same natural sample $\bar x \sim \mathcal{P} \left( {\bar x} \right)$,
where ${\small \mathcal{A}\left( \cdot \mid \bar{x} \right)}$ represents the augmentation distribution. Additionally, negative samples $\bar{x}$ are independently augmented samples, following the marginal distribution ${\small \mathcal{P}\left( x \right) = \mathbb{E}_{\bar{x}}\mathcal{A} \left( x \mid \bar{x} \right)}$.

The objective of contrastive learning is to encourage the model to map positive samples closer together in the feature space while pushing negative samples further apart.
A well-known CL objective is the InfoNCE loss \citep{oord2018representation}. 


\begin{equation}
\scalebox{0.95}{$
\begin{split}
\mathcal{L}_{\mbox{NCE}}\left(f \right) = -\mathbb{E}_{x, x_+, \left\{ {x_i^ - } \right\}_{i = 1}^M} \log \frac{{\exp \left( {f{{\left( x \right)}^T}f\left( {{x_ + }} \right)} \right)}}{{\exp \left( {f{{\left( x \right)}^T}f\left( {{x_ + }} \right)} \right) + \sum\nolimits_{i = 1}^M {\exp \left( {f{{\left( x \right)}^T}f\left( {x_i^ - } \right)} \right)} }}
\end{split}$}
\end{equation}

where $\left\{ {x_i^ - } \right\}_{i = 1}^M$ are $M$ negative samples independently drawn from $\mathcal{P}(x)$. \cite{haochen2021provable} proposed the spectral contrastive loss that is more amenable for the theoretical analysis: 
\begin{equation}
\scalebox{0.95}{$
\begin{split}
{\mathcal{L}_{{ \small \mbox{SP}}}}(f) =  - 2{\mathbb{E}_{x,x + }}f{\left( x \right)^T}f\left( {{x_ + }} \right) + {\mathbb{E}_{x,x + }}{\left( {f{{\left( x \right)}^T}f\left( {x_i^ - } \right)} \right)^2}
\end{split}$}
\end{equation}

\paragraph{Contrastive learning as matrix factorization. }
Notably, \cite{haochen2021provable}
 show that the spectral contrastive loss can be equivalently written as the following matrix factorization objective: 
\begin{equation}
\label{eq_math}
\begin{split}
\mathcal{L}_{{\small \mbox{MF}}} = {\left\| {\bar A - F{F^T}} \right\|^2}, \mbox{where} \,\, {F_{x,:}} = \sqrt {P\left( x \right)} f\left( x \right)
\end{split}
\end{equation}

Here, ${\small \bar{A} = D^{-1/2} A D^{-1/2}}$ denotes the normalized version of the co-occurrence matrix ${\small A \in \mathbb{R}_+^{N \times N}}$ of all augmented samples ${\small x \in \mathcal{X}}$ (assuming $|\mathcal{X}| = N$ for ease of exposition): ${\small \forall x, x' \in \mathcal{X}, A_{x, x'} := \mathcal{P}(x, x') = \mathbb{E}_{\bar{x}} \left[ \mathcal{A}(x|\bar{x}) \mathcal{A}(x'|\bar{x}) \right]}$.

\section{Contrastive (non-negative) factor analysis}
This section furnishes a comprehensive overview of the proposed contrastive (non-negative) factor analysis, comprising the generative model  (Sec.\ref{subsec: generative_model}) and the variational inference network (Sec.\ref{subsec: inference_network}).
It further elaborates on variational inference (Sec.\ref{subsec: elbo}) and discusses techniques for uncertainty evaluation (Sec.~\ref{subsec_stable_traing}). 
\subsection{Bayesian generative model}
\label{subsec: generative_model}
In classical FA, an observation matrix   ${\small X}$ 
is factorized into two latent variables ${\small X \sim \mbox{Dis}(\theta \, \Phi) }$, where both ${\small  \theta }$ and ${\small \Phi } $ are drawn from distributions such as Gaussian.
While FA is intuitive for feature learning, it may not be effective for learning meaningful representations with its reconstruction training objects \citep{tschannen2018recent}. 
Motivated by the connection between CL and MF, as indicated by Eq.~\ref{eq_math}, we build contrastive factor analysis (CFA) by instead modeling the co-occurrence matrix $\bar A$, which represents the relationship among positive and negative samples. 

Unlike classical FA, which focuses solely on modeling the generative process between the latent variable ${\small \theta}$ and target data ${\small X}$, CFA requires modeling the generative process among observation data samples ${\small X}$, the latent variable ${\small \theta}$, and the corresponding normalized co-occurrence matrix ${\small \bar{A}}$, as depicted in Figure.~\ref{fig_gm}.
To address this, we draw inspiration from conditional VAE \citep{sohn2015learning} to devise a modified FA. 
Specifically, given data samples and augmentations ${\small X \in \mathbb{R}^{N \times D}}$ and the corresponding normalized co-occurrence matrix ${\small \bar{A} \in \mathbb{R}{+}^{N \times N}}$, CFA factorize ${\small \bar{A} \in \mathbb{R}{+}^{N \times N}}$ under the Gaussian likelihood as follows:
\begin{equation}\label{Eq: PFA}
\begin{split}
\bar A_{i,j} \given x_{i},x_{j} \sim \mathcal{N} \left(  \theta_{i} \, \theta_{j} \right), \theta_{i} \given x_{i} \sim \mathcal{N} \left( f(x_i), \, 1 \right), \theta_{j} \given x_{j} \sim \mathcal{N} \left( f(x_j), \, 1 \right).
\end{split}
\end{equation}
where ${\small \theta_i,  \theta_j  \in \mathbb{R}_{+}^{d} }$ is the factor score
matrix, each column of which encodes the relative importance of each atom in a sample. 

Further, recent studies suggest that modeling latent features using non-negative vectors can harness the advantageous properties of non-negative matrix factorization (MF), which facilitates the acquisition of  disentangled representations \citep{lee1999learning, bengio2014representation, wang2024nonnegative}.
Hence, we draw inspiration from relevant research \citep{zhou2012beta} to develop a contrastive non-negative FA (CNFA) approach.
Specifically, the CNFA factorizes $\bar{A}$ into sparse and non-negative gamma latent variables, which can be described as follows:
\begin{equation}\label{Eq: PFA}
\begin{split}
\bar A_{i,j} \given x_{i},\,x_{j} \sim \mathcal{N} \left(  \theta_{i} \, \theta_{j} \right), \theta_{i} \given x_{i} \sim \mbox{Gamma} \left( f(x_i), 1 \right), \theta_{j} \given x_{j} \sim \mbox{Gamma} \left( f(x_j), 1 \right).
\end{split}
\end{equation}
In our formulation, the ideal scenario involves modulating the prior of the latent variables ${\small \theta_{i}}$ by the input ${\small x_{j}}$. However, we can readily relax this constraint to ensure that the latent variables are statistically independent of the input variables \citep{sohn2015learning}. Thus, for contrastive factor analysis (CFA), we simply define the data-independent prior as $f({\small x_i}) = 0$, while for contrastive non-negative factor analysis (CNFA), we set ${\small f({x}_i) = 1}$.


\subsection{Reparameterizable latent variable distributions
}
\label{subsec: inference_network}

\paragraph{Variational Gaussian posterior for CFA.} 
The Gaussian distribution $\theta \sim \mathcal{N}(\mu, \sigma^2)$ has probability density function $f\left( {\theta \given \mu ,{\sigma ^2}} \right) = \frac{1}{{\sqrt {2{\sigma ^2}\pi } }}{e^{ - {{\left( {x - \mu } \right)}^2}/2{\sigma ^2}}}$, where $\theta \in \mathbb{R}$. It is reparmeterizable as drawing 
$\theta \sim \mathcal{N}(\mu, \sigma^2)$ us equivalent to letting $\theta = g(\varepsilon) := \mu + \varepsilon \sigma , ~~\varepsilon \sim \mathcal{N}(0,1) $. The KL divergence is analysis as $\mbox{KL} \left( {\mathcal{N}\left( {{\mu _1},\sigma _1^2} \right)\,||\, \mathcal{N}\left( {{\mu _2},\sigma _2^2} \right)} \right) = \log \frac{{{\sigma _2}}}{{{\sigma _1}}} + \frac{{\sigma _1^2 + {{\left( {{\mu _1} - {\mu _2}} \right)}^2}}}{{2\sigma _2^2}}$.
As shown in Fig.~\ref{fig_vi}, the variational inference network utilize the obtained latent features ${\small h_i}$ from a deep feature extractor, such as ResNet \citep{he2015deep}, to 
construct the variational posterior:
\begin{equation}
\label{Eq_inference}
\begin{split}
    &q\left(\theta_i \given x_i \right) = 
	\mathcal{N}\left(\mu_i, \sigma^2_i \right), 
	\mu_i = f_{\mu}({h}_i),
	\sigma^2_i = \mbox{Softplus}\left(f_{\sigma}( {h}_j)\right),
\end{split}
\end{equation}where $f_{\cdot}^{(l)}(\cdot)$ denotes the neural network, and $\mbox{Softplus}$ applies $\mbox{log}(1+\mbox{exp}(\cdot))$ non-linearity to each element to ensure positive Gaussian variance parameters. 
It's evident that the CFA will reduce to matrix factorization 
as $\sigma$ of the Gaussian distribution goes to zero. Therefore, the proposed CFA can be viewed as a generalization of
vanilla contrastive learning \citep{haochen2021provable}.

\paragraph{Variational Weibull posterior for CNFA:} 
While the gamma distribution is ideal for the posterior due to its encouragement of sparsity and satisfaction of non-negativity constraints, its non-reparameterizability and incompatibility with gradient descent optimization lead to the adoption of the Weibull distribution as an approximation for the posterior of gamma latent variables.

The reason for choosing the Weibull distribution is threefold \citep{zhang2018whai}:  $\iv)$ the Weibull distribution is
similar to a gamma distribution, capable of modeling sparse,
skewed and positive distributions;
$\iv \iv)$ the Weibull distribution has a simple
reparameterization so that it is easier to optimize; That is, to sample $\theta \sim \mbox{Weibull} \left( {k,\lambda } \right)$ with probability density function $p\left( {\theta \given k,\lambda } \right) = \frac{k}{{{\lambda ^k}}}{\theta ^{k - 1}}{e^{ - {{\left( {S/\lambda } \right)}^k}}}$, where $\theta \in \mathbb{R}_{+}$, it is equivalent to letting 
$ \theta = g(\varepsilon) := \lambda {( - \ln (1 - \varepsilon ))^{1/k}},~~\varepsilon \sim \mbox{Uniform}(0,1)$.
$\iv \iv \iv)$ The KL divergence from the gamma to Weibull distributions has an analytic expression as
$\mbox{KL}\left( \mbox{Weibull}(k,\lambda) \,||\,\mbox{Gamma}(\alpha,\beta) \right)  = \frac{\gamma\alpha}{k} - \alpha \mbox{log}\lambda+ 
\mbox{log}k + \beta\lambda\Gamma(1+\frac{1}{k}) - \gamma -1 -\alpha \mbox{log}(\beta) + \mbox{log}\Gamma(\alpha).$
where $\gamma$ is the Euler-Mascheroni constant and  $\Gamma$ is the gamma function. This provides an efficient way to estimate the training
objective, which we will discuss in detail in the next subsection.
Similarly, the variational inference network employs the obtained latent features $\hv_i$ to 
construct the variational posterior:
\begin{equation}
\vspace{-2mm}
\label{Eq_inference}
\scalebox{0.95}{$
\begin{split}
    q\left(\theta_i \mid x_i \right) = \mathrm{Weibull}\left(k_i, \lambda_i \right), 
    k_i = \mathrm{Softplus}\left(f_{k}(h_i)\right), 
    \lambda_i = \mathrm{ReLU}\left(f_{\lambda}(h_i)\right)/\Gamma \left(1 + 1 /k_i \right)
\end{split}$}
\end{equation}

where $f_{\cdot}(\cdot)$ denotes the neural network, and $\mbox{Softplus}$ applies $\mbox{log}(1+\mbox{exp}(\cdot))$ non-linearity to each element to ensure positive Weibull shape and scale parameters. 
With above inference network, there are $\mathbb{E}\left( \theta  \right) = \mbox{Relu}\left( {{f_\lambda }\left( {{h_i}} \right)} \right)$.
It's evident that the CNFA will reduce to non-negative matrix factorization if the parameter $k$ of the Weibull distribution goes to infinity.
Therefore, the proposed CNFA can be viewed as a generalization of
vanilla non-negative contrastive learning \citep{wang2024nonnegative}.

\subsection{Variational inference}
\label{subsec: elbo}
For the contrastive factor analysis, given the model parameters referred to as ${\small W}$, which consist of the parameters in the generative model and inference network, marginal likelihood is defined as:
\begin{equation}
\scalebox{0.95}{$
\begin{split}
p \left( { {\bar A} \given {{W}} } \right)  = \int { { {\prod\limits_{i = 1}^N \prod\limits_{j = 1}^N {p\left( {\bar A_{i,j} \: | \: \theta_i, \theta_j}\right)} }}}  { { {\prod\limits_{j = 1}^N {p\left( {\theta _j \: | \: x _j} \right)} }}} d \theta _{j = 1}^{N} .
\end{split}$}
\end{equation}
The inference task is
to learn the parameters of the generative model and the inference network.
Similar to VAEs,
the optimization objective of contrastive factor analysis can be achieved by
maximizing the evidence lower bound (ELBO) of log-likelihood as
\begin{equation} \label{aug_ELBO}
\scalebox{0.95}{$
\begin{split}
\mathcal{L}(\bar A)   =  {\sum_{i = 1}^N \sum_{j = 1}^N {\mathbb{E}_{q(\theta_j \given x_{j})}}\left[ {\ln p\left({\bar A_{i,j}} \: | \:  \: \theta _{i}, \theta _{j} \right)} \right]}  -  {\sum_{j = 1}^N {{\mathbb{E}_{q(\theta_j \given x_j)}}\left[ {\ln \frac{{q\left(\theta _{j} \: | \: x_{j} \right)}}{{p\left(\theta _{j} \right) }}} \right]} } .
\end{split}$}
\end{equation} 
where the first term is the expected log-likelihood of the generative model, which ensures reconstruction performance, and the second term is the Kullback–Leibler (KL) divergence that constrains the variational distribution $q(\theta_j)$ to be close to its prior $p(\theta_j)$.
Unfortunately, Equation.~\ref{aug_ELBO} becomes intractable with factor analysis algorithms as the input space $X$ grows exponentially large with $N \to \infty$. To address this challenge, we reformulate the factor analysis problem (Equation.~\ref{aug_ELBO}) equivalently as a tractable contrastive learning (CL) objective that only necessitates sampling positive and negative samples from the joint and marginal distributions, respectively:
\begin{equation}
\scalebox{0.95}{$
\begin{split}
\mathcal{L} = 
& - 2{\mathbb{E}_{x,\, x + }}(\theta_{+}^T \, \theta_{+}) 
 + {\mathbb{E}_{x, \, x_{+} }}{\left( {\theta_{-}^T}\,\theta_{-} \right)^2} \\ 
&-  { {{\mathbb{E}_{q(\theta_{+} \given x_{+})}}\left[ {\ln \frac{{q\left(\theta _{+} \: | \: x_{+} \right)}}{{p\left(\theta _{+} \right) }}} \right]} } 
-  {{{\mathbb{E}_{q(\theta_{-} \given x_{-})}}\left[ {\ln \frac{{q\left(\theta _{-} \: | \: x_{-} \right)}}{{p\left(\theta _{-} \right) }}} \right]} } 
\end{split}$}
\end{equation} 
 The parameters in contrastive (non-negative) factor analysis can be directly optimized by advanced gradient algorithms, like Adam \citep{kingma2017adam}.

\paragraph{Complexity analysis} 
 Our framework is computationally and memory-efficient because only a few parameters need to be added to infer the distribution parameters. 
 The extra memory cost comes from the posterior inference network, which only adds $d \times 1$ parameters due to $\sigma_{i}^2 \in \mathbb{R}^{1}$ and $k_{i} \in \mathbb{R}^{1}$, where $d$ is the dimension of feature $h_i$.
 Meanwhile, the additional computations involve the sampling process and computing the KL term, which is of scale $O(d)$. 
 Both of above
 is inconsiderable compared to the memory and computational scale of the deep backbone neural network \citep{he2015deep}. \vspace{-2mm}
\subsection{Uncertainty-aware representation} \label{subsec_stable_traing}
To quantify the inferred latent variable uncertainty, we adopt Shannon's entropy \citep{shannon1948mathematical} of the variational posterior. Specifically, The Shannon's entropy of Gaussian variational posterior with parameters $(\mu, \sigma)$ is defined as: 
\begin{equation}
\label{eq_gaussian_entropy}
\scalebox{0.95}{$
\begin{split}
H \left( {{q_\phi }\left( {{\theta _i} \given {x_i}} \right)} \right) = \sum\limits_{m = 1}^d {\ln \left( {{\sigma _m}\sqrt {2\pi e} } \right)} 
\end{split}$}
\end{equation}. And Shannon's entropy of the Weibull variational posterior with parameters $(\lambda, k)$ is defined as: 
\begin{equation}
\label{eq_weibull_entropy}
\scalebox{0.95}{$
\begin{split}
H\left( {{q_\phi }\left( {{\theta _i} \given {x_i}} \right)} \right) = \sum\limits_{m = 1}^d {\frac{{\left( {{k_m} - 1} \right)\gamma }}{{{k_m}}} + \ln \frac{{{\lambda _m}}}{{{k_m}}} + 1} 
\end{split}$}
\end{equation}
where $\gamma$ stands for Euler's constant. 
It should be noted that the higher the entropy, the higher the uncertainty.

\section{Related work}
Our work is mainly related to factor analysis and probabilistic representation learning. Here, we introduce the related research.

\textbf{Factor analysis.} 
Before the era of deep learning, factor analysis was widely used in representation learning. And its non-negative extension \citep{blei2003latent} has achieved great progress in text analysis. Further, there is increasing interest in hierarchical factor analysis, which is aimed at exploring hierarchical data representation. 
But all this work is different from ours,  which mainly focuses on improving factor analysis with contrastive learning. 
Besides, \citep{tosh2021contrastive} proves contrastive learning is capable of revealing latent variable posterior information in FA. 
Nonetheless, our approach diverges from this earlier work as it places more emphasis on proposing a novel model that can effectively learn representations.

\textbf{Probabilistic representation learning.} 
Probabilistic representation learning has indeed made significant strides in recent years and has garnered considerable attention in the machine learning community. 
Its ability to capture uncertainty and model complex data distributions has proven immensely useful in various applications such as natural language processing \citep{zhou2016augmentable, vilnis2014word}, computer vision \citep{shi2019probabilistic, oh2018modeling, chun2021probabilistic, wang2023uncertainty}, and graph processing \citep{elinas2020variational, bojchevski2017deep}.
In contrast to the previous efforts, this paper develops a general framework for (non-negative)  contrastive factor analysis that benefits from both (non-negative) factor analysis and contrastive learning. 
Besides, there are efforts to improve contrastive learning with the Bayesian method, which is aimed at evaluating the uncertainty \citep{nakamura2023representation, park2022probabilistic, kirchhof2023probabilistic}. 
Unfortunately, most of these techniques rely on a decoder, which frequently degrades the performance of downstream tasks \citep{aitchison2021infonce}. 
On the opposite hand, our research is an encoder-only approach that can enhance downstream tasks' performance. 
\section{Experiment}
\subsection{Feature learning}\label{feature_learning}
\paragraph{Baselines \& Datasets.} 
To evaluate the expressiveness of the proposed model, we compare it with strong baselines, including SimCLR(CL) \citep{chen2020simple} and non-negative contrastive learning(NCL) \citep{wang2024nonnegative}. 
More specifically, we compare CFA with CL and CNF with NCL.
It should be noted that CFA is built on SimCLR codebase \footnote{\url{https://github.com/vturrisi/solo-learn}} with a minor modification to the hidden projector output.
We evaluate linear probing and fine-tuning performance on three benchmark datasets: CIFAR-10, CIFAR-100\citep{krizhevsky2009learning}, and Imagenet-100\citep{deng2009imagenet}. 
Apart from in-distribution evaluation, we also compare models on three out-of-distribution datasets, including stylized ImageNet\citep{geirhos2018imagenet}, ImageNet-Sketch\citep{wang2019learning}, ImageNet-C\citep{hendrycks2019benchmarking}(restricted to ImageNet-100 classes). 
For all experiments, we run 3 random trials and report their mean and standard deviation. 
All experiments are performed on
workstation equipped with a CPU i7-10700 and accelerated
by four GPU NVIDIA RTX 4090 with 24GB VRAM. 
\vspace{-2mm}
\paragraph{Experiment Setting.}\label{FL_setting}
In our experiments, Resnet-18\citep{he2015deep} is used as the backbone, and the model is trained for 200 epochs on CIFAR-10, CIFAR-100, and 100 epochs for ImageNet-100. 

\vspace{-2mm}
\paragraph{In distribution experiment results.} The experiment results are  exhibited in Table~\ref{in-distribution}. 
Firstly, compared to the FA model, CFA has achieved a significant improvement, primarily attributed to the effective contrastive learning training strategies and corresponding expressive deep neural networks.
Secondly, acting as the Bayesian extension of contrastive learning (CL) and non-negative contrastive learning (NCL), respectively, CFA and CNFA demonstrate varied degrees of improvement across different datasets. This result indicates the efficacy of stochastic latent variables in capturing complex data dependencies and enhancing overall performance \citep{fan2020bayesian}.
Finally, when compared with CFA, CNFA consistently exhibits superior performance across all datasets, as NCL performs better than CL. This improvement may stem from the ability of non-negative stochastic latent variables to learn  disentangled representations \citep{bengio2014representation, wang2024nonnegative}.

\textbf{Out-of-distribution experiment results.} Apart from the expressiveness evaluation, we further evaluate the model's robustness on the out-of-distribution dataset, and the results are depicted in Table~\ref{OOD}. 
Firstly, compared to deterministic contrastive learning methods, CFA and CNFA show significant improvement in out-of-distribution generalization downstream tasks.
The improved OOD performance can be partially attributed to the generalization introduced by the stochastic latent variables, which allows the model to handle unseen and diverse data better \citep{kingma2013auto}.
By inheriting the advantages of non-negative factor analysis, CNFA can learn disentangled representation \citep{wang2024nonnegative} and exhibit better transfer ability compared with CFA.
\begin{table}[t]
\renewcommand{\arraystretch}{1.25}
\caption{In-distribution evaluation on three benchmark datasets. The best results are highlighted in bold, while the second-best results are underlined.}
\centering
\scalebox{0.9}{
\begin{tabular}{lccccccccc}
\toprule
\multicolumn{1}{c}{Methods}&\multicolumn{3}{c}{CIFAR10}& \multicolumn{4}{c}{CIFAR100} &\multicolumn{2}{c}{ImagNet-100}\\
\midrule
&&LP&FT&&LP&FT&&LP&FT\\
\midrule
 FA &&35.4 $\pm$ 0.6 &52.3 $\pm$ 0.5 && - & - && - & - \\
\midrule
 CL &&87.6 $\pm$ 0.2 &92.3 $\pm$ 0.1 &&58.6 $\pm$ 0.2 &72.6 $\pm$ 0.1 &&68.7 $\pm$ 0.3 &77.3 $\pm$ 0.5\\
\textbf{CFA} &&\underline{88.0 $\pm$ 0.2} &\underline{92.6 $\pm$ 0.1} &&\underline{60.9 $\pm$ 0.2} &\underline{73.0 $\pm$ 0.2} && \underline{69.6 $\pm$ 0.4} &\underline{78.2 $\pm$ 0.3}\\
 \midrule
 NCL&&87.8 $\pm$ 0.2 &92.6 $\pm$ 0.1 &&59.7 $\pm$ 0.4 &73.0 $\pm$ 0.3 && 69.4 $\pm$ 0.3 &79.2 $\pm$ 0.4\\
 \textbf{CNFA} &&\textbf{88.2 $\pm$ 0.2} &\textbf{92.8 $\pm$ 0.2} &&\textbf{61.1 $\pm$ 0.1}&\textbf{73.3 $\pm$ 0.2} &&\textbf{70.0 $\pm$ 0.2} &\textbf{80.2 $\pm$ 0.4}\\
\bottomrule
\end{tabular}}
\label{in-distribution}
\end{table}
\begin{table}[t]
\renewcommand{\arraystretch}{1.25}
    \caption{Out-of-distribution transferability on three benchmark datasets. The best results are highlighted in bold, while the second-best results are underlined.}
    \centering
    \scalebox{0.9}{
    \begin{tabular}{lccccccc}
    \toprule
    Method&\multicolumn{2}{c}{Stylized}
         &\multicolumn{3}{c}{Corruption} &\multicolumn{2}{c}{Sketch}\\
    \midrule
    CL &\multicolumn{2}{c}{12.3 $\pm$ 0.4} &\multicolumn{3}{c}{34.5 $\pm$ 0.2} &\multicolumn{2}{c}{27.1 $\pm$ 0.1} \\
    \textbf{CFA} &\multicolumn{2}{c}{\underline{13.3 $\pm$ 0.2} }&\multicolumn{3}{c}{\underline{36.9 $\pm$ 0.2}}& \multicolumn{2}{c}{\underline{28.0 $\pm$ 0.3}} \\
    \midrule
    NCL &\multicolumn{2}{c}{12.7 $\pm$ 0.4} & \multicolumn{3}{c}{36.1 $\pm$ 0.3}  & \multicolumn{2}{c}{28.0 $\pm$ 0.2 }\\
    \textbf{CNFA} &\multicolumn{2}{c}{\textbf{14.4 $\pm$ 0.3}} &\multicolumn{3}{c}{\textbf{37.8 $\pm$ 0.3} }& \multicolumn{2}{c}{\textbf{28.3 $\pm$ 0.2}} \\
    \bottomrule
    \end{tabular}}
    \vspace{-3mm}
    \label{OOD}
\end{table}

\subsection{Uncertainty evaluation}\label{uncerainty_learning}
In addition to their enhanced expressiveness and robustness on out-of-distribution (OOD) data, another notable strength of CFA and CNFA is their ability to capture data uncertainty.

\textbf{Evaluation Metric.} 
 With the entropy of the testing result and a given entropy threshold, we can determine whether the model is certain or uncertain about one prediction.  
 To evaluate the
uncertainty estimates, we use Patch Accuracy vs Patch Uncertainty (PAvPU) \citep{mukhoti2019evaluating}, which is
defined as PAvPU $ = (n_{ac} + n_{iu})/(n_{ac} + n_{au} + n_{ic} + n_{iu})\label{pavpu_top1}$, where $n_{ac}, n_{au}, n_{ic}, n_{iu}$  represent the numbers of accurate and certain, accurate and uncertain, inaccurate and certain, and inaccurate and uncertain samples, respectively. Compared to Top-1 Accuracy$ = (n_{ac} + n_{au})/(n_{ac} + n_{au} + n_{ic} + n_{iu})$,
 PAvPU metric would be higher if the model tends to
generate accurate predictions with low uncertainty and inaccurate predictions with high uncertainty leading to a more trustworthy algorithm.

\textbf{Evaluation Setting.}
Firstly, utilizing Equation~\ref{eq_gaussian_entropy} and Equation~\ref{eq_weibull_entropy}, we calculate the information entropy of each sample in the test set. Subsequently, we select the M samples with the highest information entropy as uncertain samples, while the remaining samples are considered definite. In practice, we conducted experiments with different values of M, specifically selecting 100, 200, and 400 samples as uncertain samples to compute the test indices.


\textbf{Experiment results:}
Table.~\ref{pavpu_results}  show the PAvPU results of CFA and CNFA on ImageNet-100. CFA and CNFA achieves better performance on PAvPU metrics compared to Top-1 accuracy. 
This suggests that CFA and CNFA are effective not only in making correct predictions but also in identifying when it is likely to be wrong, which is crucial for applications requiring high reliability. 
\begin{figure*}[t]
\centering
\subfigure[Top-20 images have the highest uncertainty]
{
\includegraphics[width=0.4\linewidth]{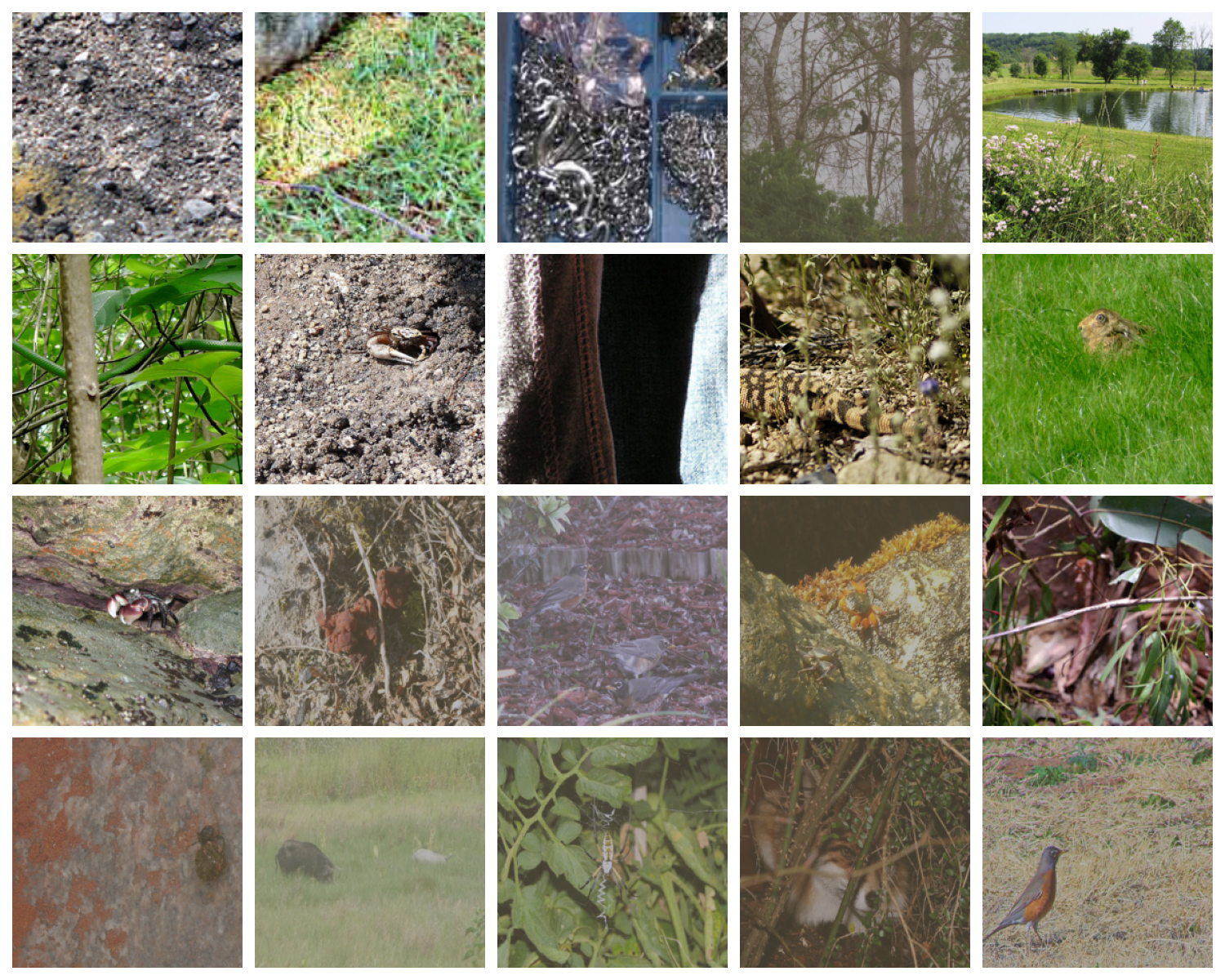}
\label{uncertainty_vis}
}
\quad \quad
\subfigure[Top-20 images have the lowest uncertainty]{
\includegraphics[width=0.4\linewidth]{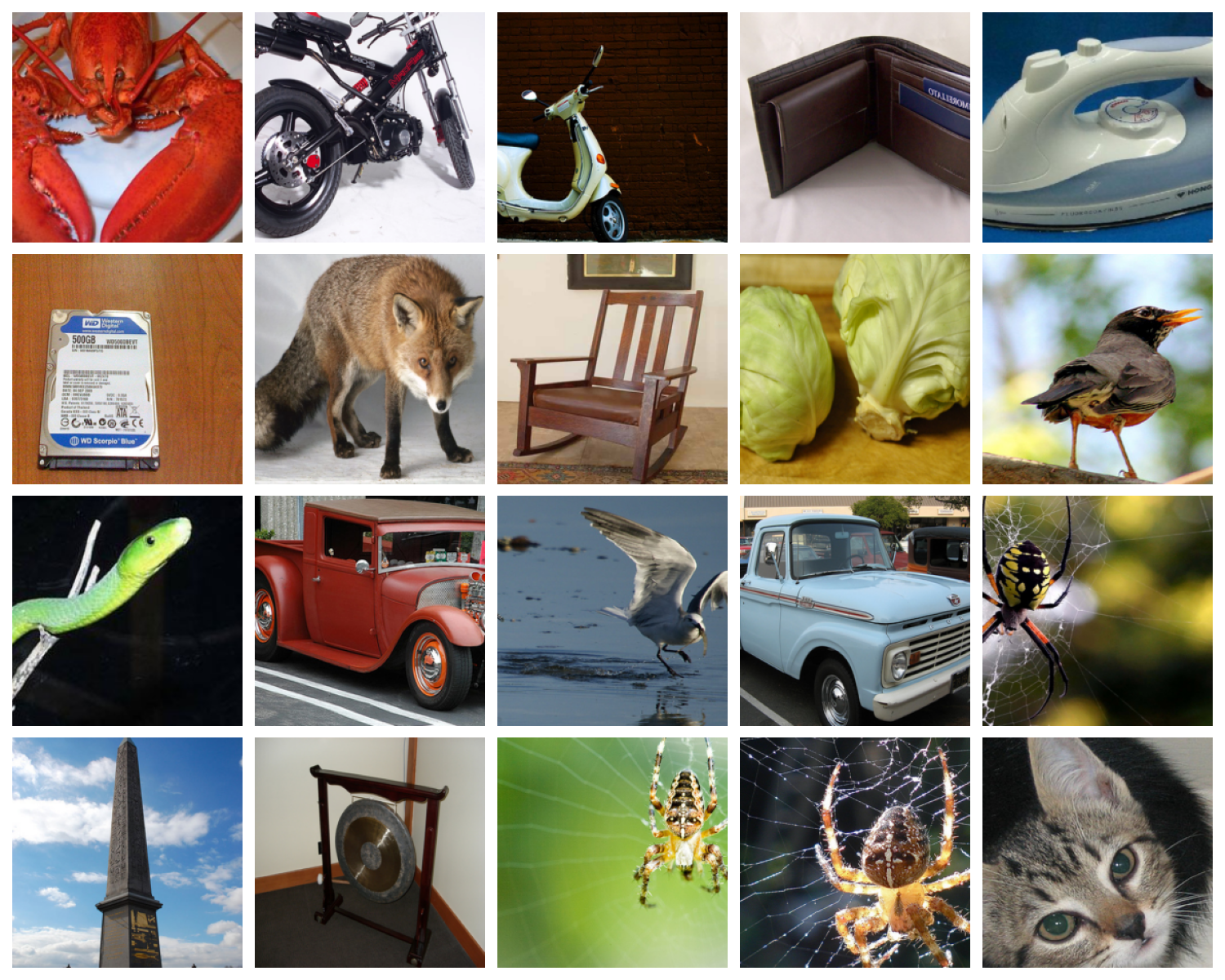}
\label{certainty_vis}
} \vspace{-2mm}
\caption{
We present 20 images on ImageNet-100 with the highest and lowest uncertainty predicted by CNFA. 
Figure.~\ref{uncertainty_vis} visually indicates that images with noisy backgrounds exhibit higher levels of uncertainty. 
In contrast, Figure.~ \ref{certainty_vis} shows that a clean background with high contrast is prone to displaying low uncertainty. 
This clear distinction highlights the effectiveness of CNFA in identifying the factors contributing to uncertainty in image predictions. 
} \label{vis_all}
\end{figure*} 
\begin{figure*}[!]
\centering
\subfigure[CFA]
{
\includegraphics[width=0.4\linewidth]{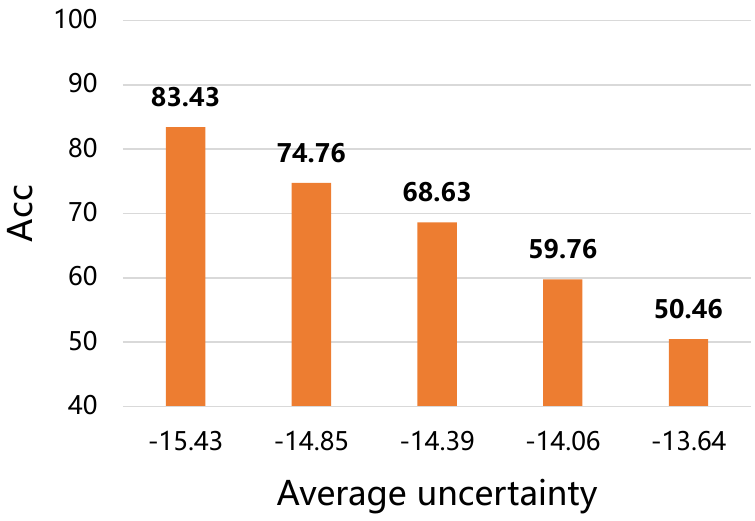}
}
\quad \quad \quad
\subfigure[CNFA]{
\includegraphics[width=0.4\linewidth]{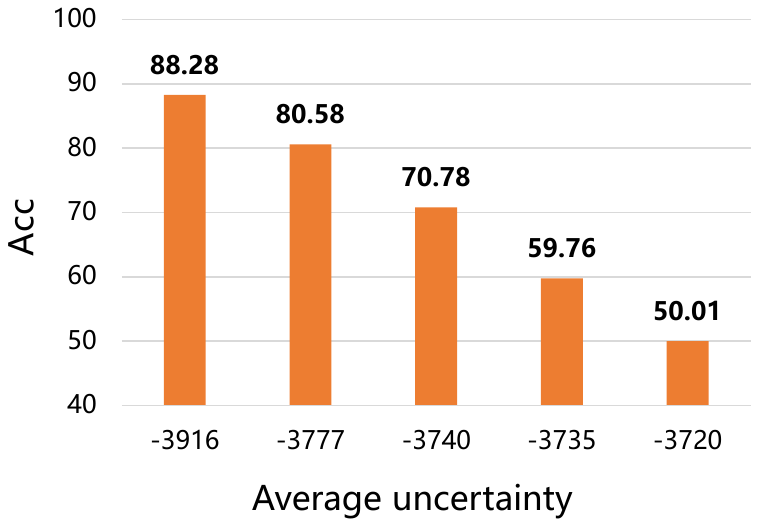}
} \vspace{-2mm}
\caption{
Utilizing the entropy of samples within the test set, we organized them into five subsets based on their respective entropy levels. Subsequently, we plot histograms, illustrating the average entropy and accuracy for each of these subsets.
The histograms clearly depict a negative correlation between entropy and accuracy: as entropy increases (uncertainty becomes greater), accuracy decreases.
} \label{vis_acc_uncetainty}
\vspace{-3mm}
\end{figure*}

\textbf{Qualitative analysis:}
To further validate the efficacy of the proposed model in uncertain evaluation, we conducted qualitative analysis. Specifically, we selected 20 images with the highest and lowest uncertainty predicted by CNFA from the ImageNet-100 dataset, and the experimental results are illustrated in Figure~\ref{vis_all}.
Figure.~\ref{uncertainty_vis} visually indicates that images with noisy backgrounds exhibit higher levels of uncertainty. 
In contrast, Figure.~ \ref{certainty_vis} shows that a clean background alongside high contrast is prone to displaying low uncertainty. 
This clear distinction highlights the effectiveness of CNFA in identifying the factors contributing to uncertainty in image predictions. 

\textbf{Ablation study:}
To further investigate the relationship between prediction uncertainty and downstream performance, we conducted an ablation study. Specifically, we sorted the samples in the test set based on their evaluation uncertainty and divided the set into five subsets accordingly. Subsequently, we calculated the average accuracy and corresponding average uncertainty for each subset. We then plotted histograms depicting the average uncertainty and accuracy for these five subsets, as illustrated in Figure.~\ref{vis_acc_uncetainty}. The histogram clearly indicates a negative correlation between uncertainty and accuracy: higher uncertainty correlates with lower accuracy. It's worth noting that these findings are supported by qualitative analysis, where higher uncertainty suggests noisier and more challenging-to-classify data.

\begin{table}[h]
    \centering
    \caption{PAvPU results on ImageNet-100 dataset, where @M denotes the number of uncertain samples. The PAvPU metric provides a more reliable measure of classification accuracy, as previously discussed in \ref{pavpu_top1}. CFA-like methods exhibit higher PAvPU values compared to Top-1 accuracy, indicating that our proposed methods are effective in making accurate predictions and identifying highly uncertain samples.}
    \label{pavpu_results}
    
    (a) Linear Probing Evaluation \\
    \scalebox{0.9}{
    \begin{tabular}{lcccc}
    \toprule
         Methods & Top-1 Acc & PAvPU@100 & PAvPU@200 & PAvPU@400 \\
         \midrule
         \textbf{CFA} & 69.6 $\pm$ 0.4 & 69.6 $\pm$ 0.3 & \textbf{69.8 $\pm$ 0.2} & 69.7 $\pm$ 0.2 \\
         \textbf{CNFA} & 70.0 $\pm$ 0.2 & 70.3 $\pm$ 0.3 & \textbf{70.8 $\pm$ 0.2} & 70.6 $\pm$ 0.2 \\
    \bottomrule
    \end{tabular}}
    
    \vspace{3mm}
    (b) Fine-Tuning Evaluation \\
    \scalebox{0.9}{
    \begin{tabular}{lcccc}
    \toprule
         Methods & Top-1 Acc & PAvPU@100 & PAvPU@200 & PAvPU@400 \\
         \midrule
         \textbf{CFA} & 78.2 $\pm$ 0.3 & \textbf{78.5 $\pm$ 0.3} & 77.2 $\pm$ 0.2 & 77.9 $\pm$ 0.2 \\
         \textbf{CNFA} & 80.2 $\pm$ 0.4 & 80.2 $\pm$ 0.2 & 80.2 $\pm$ 0.2 & \textbf{80.6 $\pm$ 0.2} \\
    \bottomrule
    \end{tabular}}
\end{table}

\begin{table}[!]
\renewcommand{\arraystretch}{1.25}
\centering
\vspace{-2mm}
\caption{Feature disentanglement score on ImageNet-100, where @k denotes the top-k dimensions. Values are scaled by $10^2$.} 
\scalebox{0.9}{
\begin{tabular}{lcccc}
\toprule
Methods  & SEPIN@10  & SEPIN@100 & SEPIN@1000 &SEPIN@all \\ 
\midrule
 CL & 0.79 $\pm$ 0.02 & 0.69 $\pm$ 0.01 & 0.54 $\pm$
 0.01 &0.47 $\pm$ 0.01 \\
  NCL  & 5.93 $\pm$ 0.12 & 3.84 $\pm$ 0.04 &0.94 $\pm$
 0.02 &0.48 $\pm$ 0.01 \\
  \textbf{CNFA}  & \textbf{6.06 $\pm$ 0.08} & \textbf{4.32 $\pm$ 0.04} & \textbf{0.95 $\pm$
 0.01} &\textbf{0.48 $\pm$ 0.01} \\
 \bottomrule
\end{tabular}\label{disentangle}}
\vspace{-2mm}
\end{table}

\vspace{-2mm}
 
\subsection{Disentanglement representation learning}
Learning a robust representation involves extracting explanatory factors that are sparse, disentangled, and semantically meaningful \citep{bengio2014representation}. By incorporating non-negative constraints, NCL can learn disentangled features \citep{wang2024nonnegative}. On the other hand, the Gamma distribution promotes sparsity due to its probability density function, which helps filter out irrelevant features and focus on the most significant ones. Thus, we evaluate CFA's performance in disentangled representation learning.
\paragraph{Evaluation Metric.} 
Many disentanglement metrics like MIG \citep{chen2019isolating} are supervised and require ground-truth factors which are often unavailable in practice. Consequently, current disentanglement evaluation is often limited to synthetic data\citep{carbonneau2022measuring}. To validate the disentanglement on real-world data, we adopt an unsupervised disentanglement metric SEPIN@k\citep{do2021theory}. SEPIN@k measures how each feature $f_i(x)$ is disentangled from others $f_{\neq i}(x)$ by computing their conditional mutual information with the top k features, i.e., SEPIN@k$= \frac{1}{k}\sum_{i=1}^{k}I(x, f_{r_i}(x)|f_{\neq r_i}(x))$, which are estimated with InfoNCE lower bound \citep{oord2018representation} implemented following \citep{wang2024nonnegative}.

\paragraph{Experiment Result.} Table~\ref{disentangle} demonstrates that CNFA features show better disentanglement than CL and NCL in all top-k dimensions. The advantage is larger by considering the top features, since learned features may contain noisy dimensions \citep{wang2024nonnegative,yu2003feature}. Incorporating proper prior distribution (e.g., Gamma distribution) will effectively mitigate this problem. Thanks to the inherent properties of the Gamma distribution, CNFA tends to produce disentangled features, making it more effective in distinguishing the underlying factors of variation within the data. In conclusion, CNFA achieves better feature disentanglement on real-world data and exhibits exceptional performance in estimating uncertainty.


\section{Conclusion}
In conclusion, this paper introduces a novel framework called Contrastive Factor Analysis (CFA), which bridges the gap between traditional Factor Analysis (FA) and the powerful technique of contrastive learning. By leveraging the mathematical equivalence between contrastive learning and matrix factorization, CFA capitalizes on the strengths of both methods, offering enhanced expressiveness, robustness, and interpretability.
Further, by extending CFA to a non-negative version (CNFA), we exploit the interpretability properties of non-negative factor analysis, enabling the learning of disentangled representations. Through extensive experimental validation, our methodology demonstrates efficacy across various dimensions, including expressiveness, robustness, interpretability, and accurate uncertainty estimation.
In essence, CFA and CNFA represent promising approaches that amalgamate the insights from factor analysis and contrastive learning, paving the way for more effective unsupervised representational learning in the deep learning era.

\bibliography{neurips_2024}
\bibliographystyle{unsrtnat}
\newpage

\end{document}